\documentclass[journal]{IEEEtran}
\usepackage{amsmath,amsfonts}
\usepackage{algorithmic}
\usepackage{algorithm}
\usepackage{array}
\usepackage[caption=false,font=normalsize,labelfont=sf,textfont=sf]{subfig}
\usepackage{textcomp}
\usepackage{stfloats}
\usepackage{url}
\usepackage{verbatim}
\usepackage{graphicx}
\usepackage{cite}
\usepackage{multirow}
\usepackage{tcolorbox}
\begin{document}

\title{Learning-enhanced electronic skin for tactile sensing on deformable surface based on electrical impedance tomography}

\author{Huazhi Dong,\IEEEmembership{ Student Member,IEEE},
Xiaopeng Wu,\IEEEmembership{ Student Member,IEEE},
Delin Hu,
Zhe Liu,\IEEEmembership{ Graduate Student Member,IEEE},
Francesco Giorgio-Serchi,\IEEEmembership{ Member,IEEE}, 
and Yunjie Yang,\IEEEmembership{ Senior Member,IEEE}
\thanks{This work was supported in part by the European Research Council Starting Grant under Grant no.101165927 (Project SELECT).}
\thanks{Huazhi Dong, Xiaopeng Wu, Delin Hu, Zhe Liu and Yunjie Yang are with the Institute for Imaging, Data and Communications, School of Engineering, The University of Edinburgh, EH9 3BF Edinburgh, U.K. (e-mail: huazhi.dong@ed.ac.uk; xiaopeng.wu@ed.ac.uk; delin.hu@ed.ac.uk; zz.liu@ed.ac.uk; y.yang@ed.ac.uk).}
\thanks{Francesco Giorgio-Serchi is with the Institute for Integrated Micro and Nano Systems, School of Engineering, The University of Edinburgh, EH8 9YL Edinburgh, U.K. (e-mail: F.Giorgio-Serchi@ed.ac.uk).}}

\markboth{Journal of \LaTeX\ Class Files,~Vol.~14, No.~8, August~2021}%
{Shell \MakeLowercase{\textit{et al.}}: A Sample Article Using IEEEtran.cls for IEEE Journals}


\maketitle

\begin{abstract}
Electrical Impedance Tomography (EIT)-based tactile sensors offer cost-effective and scalable solutions for robotic sensing, especially promising for soft robots. However a major issue of EIT-based tactile sensors when applied in highly deformable objects is their performance degradation due to surface deformations. This limitation stems from their inherent sensitivity to strain, which is particularly exacerbated in soft bodies, thus requiring dedicated data interpretation to disentangle the parameter being measured and the signal deriving from shape changes. This has largely limited their practical implementations. This paper presents a machine learning-assisted tactile sensing approach to address this challenge by tracking surface deformations and segregating this contribution in the signal readout during tactile sensing. We first capture the deformations of the target object, followed by tactile reconstruction using a deep learning model specifically designed to process and fuse EIT data and deformation information. Validations using numerical simulations achieved high correlation coefficients (0.9660 - 0.9999), peak signal-to-noise ratios (28.7221 - 55.5264 dB) and low relative image errors (0.0107 - 0.0805). Experimental validations, using a hydrogel-based EIT e-skin under various deformation scenarios, further demonstrated the effectiveness of the proposed approach in real-world settings. The findings could underpin enhanced tactile interaction in soft and highly deformable robotic applications.
\end{abstract}


\section{Introduction}
\label{sec:introduction}
\IEEEPARstart{T}{he} research on electronic skins (e-skins) for robotics is largely inspired by the sensory capabilities of the human skin. Human skin is notably equipped with a dense network of mechanoreceptors, which allow for the detection of a wide variety of tactile stimuli thus facilitating comprehensive tactile perception through cognitive processing \cite{Loomis1986, Booth2018}. A key goal in robotics is to emulate these tactile sensing capabilities across a robot's body surface. Achieving this objective will enable robots to perceive and interact with their surroundings more naturally and robustly, and contribute to developing safer and more intelligent human-machine interactions \cite{Yang2023, Cheng2019}.

Developing a robotic skin capable of conforming to the complex and varied shapes of the robot's surface poses a significant challenge. The predominant approaches involve the use of flexible artificial skins that consist of an array of discrete sensing elements \cite{SilveraTawil2015, Sundaram2019}. However, as the sensing area increases, so does the number of sensors, leading to complex wiring, interface and higher costs. Consequently, simplifying sensor configuration while managing expenses has become a priority in manufacturing robotic skins \cite{Dahiya2013_2}. One promising strategy involves array indexing, utilizing orthogonal stretchable electrode lines to constitute an array of sensing elements \cite{Boutry2018, Won2019}. This strategy significantly reduces the number of electrodes required, yet it still demands many wires to encompass a large area with maintained high spatial resolution.

Electrical Impedance Tomography (EIT) has emerged as a promising approach for comprehensive robotic skin coverage with sparse boundary electrode configurations \cite{Park2021, Tawil2011}. EIT reconstructs conductivity distribution within a Region Of Interest (ROI) by injecting a known current and measuring the induced voltage at the boundary of the ROI. Several previous studies have demonstrated the potential of EIT-based flexible robotic skins. Kato \textit{et al.} \cite{Kato2007} created a single layer of pressure-sensitive skin based on EIT by blending rubber with conductive carbon particles. Nagakubo \textit{et al.} \cite{Nagakubo2007} made a single layer of high-tension pressure-sensitive skin based on EIT by spraying a conductive water-based carbon coating on the surface of a common fabric. Liu \textit{et al.} \cite{Liu2015} demonstrated that EIT-based e-skin offers advantages in terms of safety, affordability, and ease of manufacture, making it a potential solution for the widespread tactile perception challenge in robotics. However, existing methods are generally tested on fixed sensing surfaces under static conditions, disregarding the dynamic deformations that often occur in real-world scenarios. In response, Alirezaei \textit{et al.} \cite{Alirezaei2009} developed a distributed tactile sensor capable of delivering consistent measurements even under dynamic and substantial stretching. Li \textit{et al.} \cite{Li2024} reported a stretchable e-skin based on laser-induced graphene and liquid metal to create stretchable resistive tactile pressure sensors in an elastic eco-bending polymer. However, these approaches reconstruct tactile maps on the undeformed morphology instead of incorporating the true deformed model, which leads to errors and compromises in reconstruction accuracy.
\begin{figure}[t]
\centerline{\includegraphics[scale=0.65]{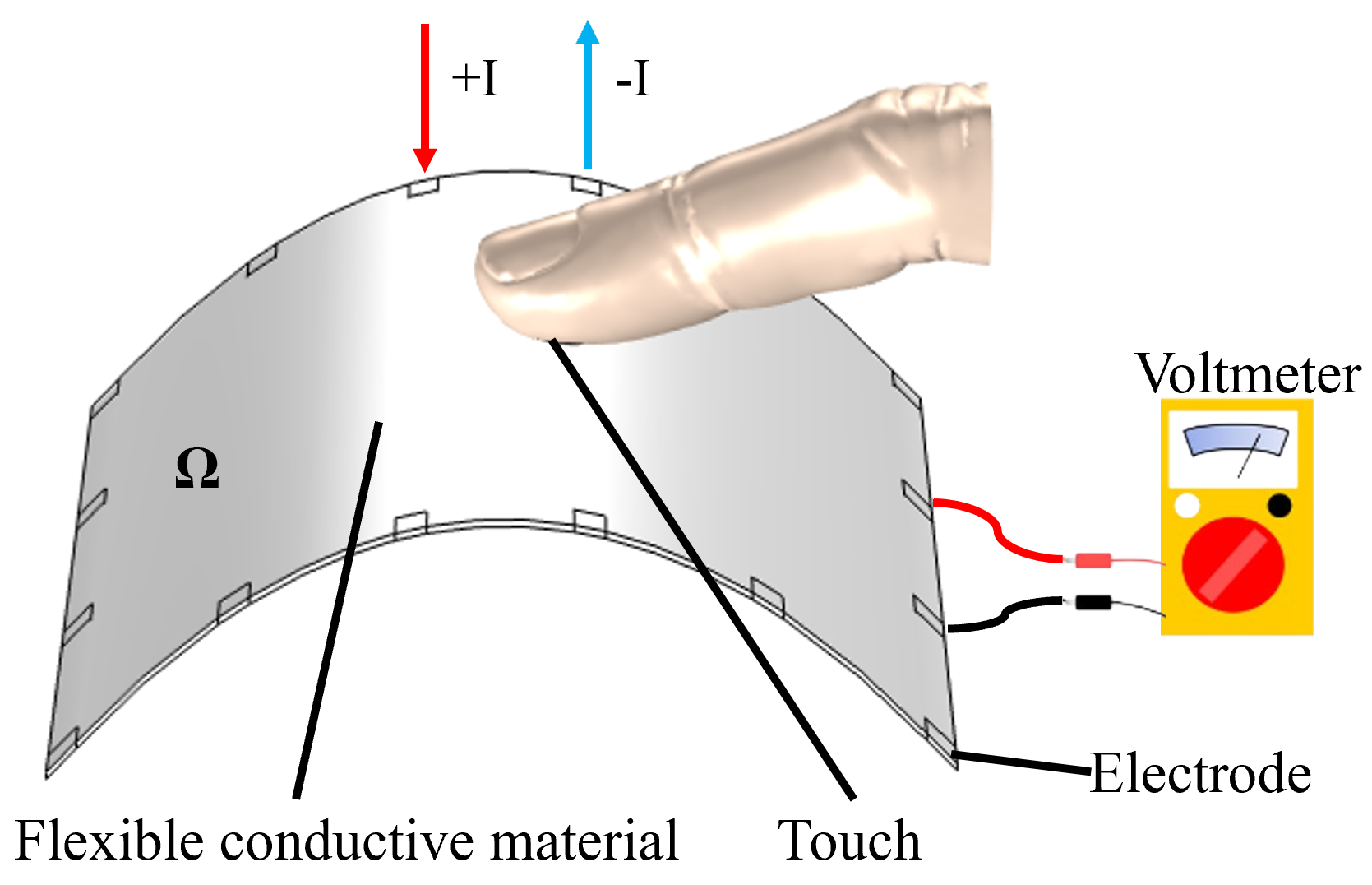}}
\caption{Schematic illustration of EIT-based tactile sensing.}
\label{Fig1}
\end{figure}
\begin{figure*}[t]
\centering
\includegraphics[scale=0.35]{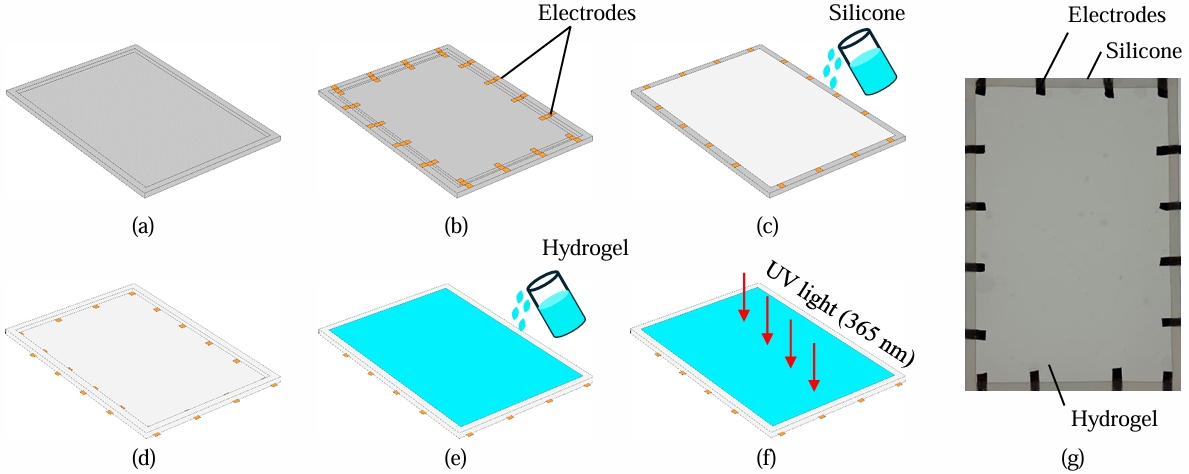}
\caption{Fabrication process of the EIT-based tactile e-skin. (a) 3D printed mould. (b) Deploy electrodes on the mould. (c) Eco-flex was poured into the mould. (d) Cure at room temperature for 4 hours and release the mould. (e) The pre-gel solution of hydrogel was poured onto the cured silicone. (f) The hydrogel was polymerized by exposing it to UV light (365 nm) for 2 hours. (g) The fabricated EIT-based tactile sensor.}
\label{Fig2}
\end{figure*}

Here, we introduce a learning-assisted tactile reconstruction method, incorporating sensor deformation information into  EIT-based tactile reconstructions and ultimately enabling tactile estimation along with the true shape of the deformed sensor. Our approach first captures point cloud data that describes the deformations of the sensor via 3D scanning and then performs the reconstruction of tactile interactions on the deformed surface using a Voltage and Deformation to Tactile (VD2T) model to process and fused EIT measurements and point cloud information. Additionally, we leverage hydrogel to construct the real-world EIT-based tactile sensor thanks to their flexibility, bio-compatibility and favourable conductivity to constitute the sensing layer. Our contributions are as follows:
\begin{itemize}
\item We propose an EIT-based tactile sensing approach that incorporates deformations in the EIT forward model, resulting in high-quality tactile reconstructions on deformed surfaces.
\item We introduce the VD2T model, a first-in-its-kind approach designed to integrate and analyze simultaneously both EIT measurements and 3D point cloud data depicting the sensor shape. This model addresses the challenges brought by sensor shape variations in tactile reconstruction, ensuring accurate tactile imaging across various surface deformations.
\item We develop a flexible EIT-based tactile sensing e-skin made of hydrogel and demonstrate and validate the proposed approach with real-world experiments.
\end{itemize}


\section{EIT for Tactile Sensing}
 Fig. \ref{Fig1} illustrates the sensing principle of EIT-based tactile sensors. On a flexible, conductive substrate, it sequentially injects current into the boundary electrodes and collects the voltage measurements $ \mathbf{V} \in \mathbb{R}^m $ using a predetermined sensing strategy. The EIT-reconstruction problem in tactile sensing is to estimate the conductivity distribution $ \boldsymbol{\sigma} \in \mathbb{R}^n $ induced by touch in the sensing region $\boldsymbol{\Omega}$ through the voltage measurements. Generally, EIT involves two sub-problems: the forward and inverse problems. The forward problem describes the relationship between $ \mathbf{V}\in \mathbb{R}^{m} $ and $ \boldsymbol{\sigma}\in \mathbb{R}^{n}$:
\begin{equation}
\mathbf{V}=F(\boldsymbol{\sigma})
\label{eq1}
\end{equation}
where $F$ represents the nonlinear forward mapping and the superscript $ n $ identifies the number of points within the point cloud representing the deformed surface. On the contrary, the inverse problem aims to estimate the conductivity within the ROI taking voltage measurements as the input. EIT-based tactile sensing adopts time-difference imaging \cite{Kuen2009} by linearizing \eqref{eq1} as:
\begin{equation}
\Delta \mathbf{V}=\boldsymbol{J} \Delta \boldsymbol{\sigma}
\label{eq2}
\end{equation}
where $\boldsymbol{J} \in \mathbb{R}^{m \times n}$ is the Jacobian matrix; $\Delta \mathbf{V}$ is the voltage change and $\Delta \boldsymbol{\sigma}$ is the conductivity change with respect to a reference. We take the normalized format of \eqref{eq2} for improved performance, where $\Delta \mathbf{V}= \pm\left(\mathbf{V}_t-\mathbf{V}_{t_0}\right) \oslash \mathbf{V}_{t_0}$ and $\Delta \boldsymbol{\sigma}=\mp\left(\boldsymbol{\sigma}_t-\boldsymbol{\sigma}_{t_0}\right) \oslash \boldsymbol{\sigma}_{t_0}$. Here, $t$ is the probing time and ${t_0}$ is the reference time point. $\oslash$ denotes the Hadamard division. Under the model-based framework, the EIT inverse problem targeting to estimate conductivity change induced by touch can generally be formulated as:
\begin{equation}
\Delta \hat{\sigma} =\arg \min _{\Delta \boldsymbol{\sigma}}\|\boldsymbol{J} \Delta \boldsymbol{\sigma}-\Delta \mathbf{V}\|_2^2+\tau R(\Delta \boldsymbol{\sigma})
\label{eq3}
\end{equation}
where $R$ represents the regularizer that incorporates prior information and $\tau>0$ is the regularization factor. 

Ultimately, the learning-based framework aims to find an inverse mapping operator $F^{-1}$ via data driven approaches:
\begin{equation}
\Delta \hat{\sigma}=F^{-1}(\Delta V)
\label{eq4}
\end{equation}

\begin{figure*}[t]
\centering
\includegraphics[width=\textwidth]{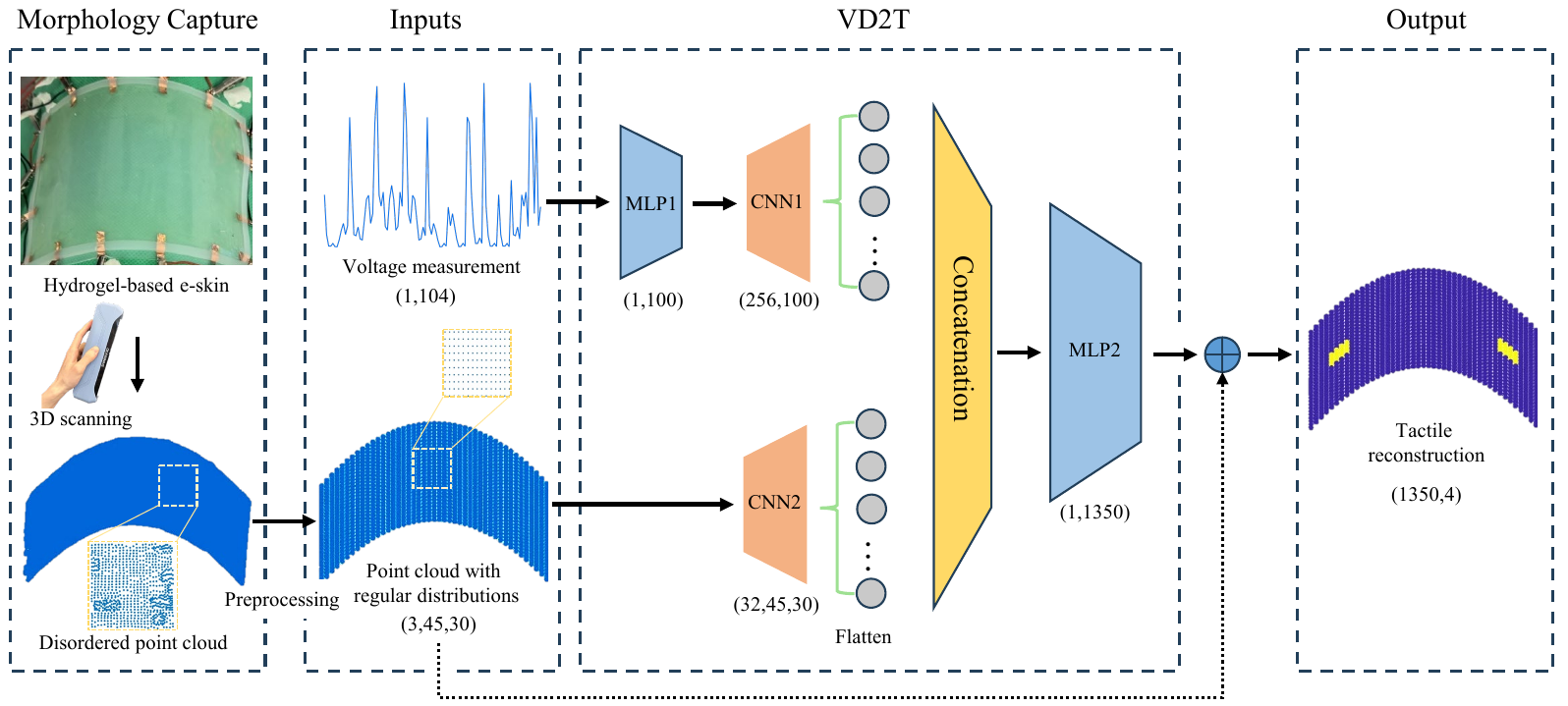}
\caption{The architecture of the proposed VD2T model.}
\label{Fig3}
\end{figure*}
\section{Hydrogel-based EIT Tactile E-skin}
We develop a hydrogel-based EIT tactile e-skin for deformable tactile sensing motivated by recent studies that demonstrated hydrogel's effectiveness for this field of application. For instance, Zhu \textit{et al.} \cite{Zhu2020} developed a hydrogel-based EIT sensor compliant with the tissue surface, which provided continuous spatial mapping of deformations. Park \textit{et al.} \cite{Park2022} reported a multi-layered hydrogel-based EIT sensor for recovering tactile information. Owing to their transparency, stretchability and ability to retain favourable conductivity \cite{Keplinger2013}, we adopted hydrogel as the conductive layer in our sensor.

Fig. \ref{Fig2} shows the fabrication process and design of the EIT-based tactile e-skin. It consists of a silicone layer, a hydrogel sensing layer and 16 copper electrodes evenly distributed along the boundaries. The overall dimensions are 160 $\times$ 110 $\times$ 3 mm$^{3}$. The conductive layer, positioned within the sensor, measures 150 $\times$ 100 $\times$ 2 mm$^{3}$. The contact area of each electrode with the conductive layer is 4 $\times$ 4 mm$^{2}$. Silicone elastomer was prepared by mixing parts A and B of Ecoflex 00-30 (Smooth-On Inc.) with a ratio of 1:1 and then cast in a 3D printed mould and cured at room temperature for 4 hours. The cured silicone is removed from the mould poured into the pre-gel hydrogel solution and irradiated under ultraviolet light (365 nm, CL-508) for 2 hours.

We prepared the hydrogel following the protocol outlined in \cite{Zhu2020}. First, 21.5 weight \% (wt\%) lithium chloride (Sigma-Aldrich) and 8 wt\% acrylamide monomer (Sigma-Aldrich) were dissolved in 32.27 wt\% deionized water. The mixture was stirred magnetically for 2 hours to ensure complete dissolution. Subsequently, 37 wt\% ethylene glycol (Sigma-Aldrich) was added, and the solution was magnetically stirred for another 2 hours. Next, 1 wt\% polyacrylamide (Sigma-Aldrich) was introduced, and the solution was magnetically stirred overnight. Following this, 0.15 wt\% N, N’-methylenebisacrylamide (Sigma-Aldrich) and 0.08 wt\% 2-hydroxy-2-methylpropiophenone (Sigma-Aldrich) were included in the mixture and stirred for 2 hours in a location shielded from light. It is worth noting that the magnetic stirring in previous steps should be performed at 60 ${ }^{\circ} \mathrm{C}$ at 700 rpm.

\section{Deformable Tactile Sensing Framework}
\subsection{Tactile Reconstruction under Deformed Surfaces}
Fig. \ref{Fig3} illustrates the proposed tactile reconstruction framework on deformed surfaces, which consists of three main steps:
\begin{enumerate}
\item 3D scanning: The first step involves conducting a high-precision 3D scan of the e-skin (Shining 3D EINSTAR). This process captures point cloud data that accurately represents the surface of the e-skin \cite{Dong2024}. Its resolution is 0.1 mm, ensuring detailed and reliable surface representation. It is noteworthy that the point cloud data is obtained before touch interaction occurs and does not include information on the touch areas, which only provides the geometric information of the deformed sensor surface.
\item Point cloud pre-processing: Radial Basis Function (RBF) interpolation \cite{Franke1982} is employed to create a smooth surface of the original point cloud, effectively mitigating irregularities and filling gaps. Then, the coordinates are reorganized and outliers are removed to enhance data consistency. Additional processing includes resampling to a suitable grid and rotational adjustments for optimal alignment within the coordinate system. The pre-processing can improve the uniformity of the point cloud and transform it from a disordered state into a regularly distributed structure.
\item Tactile reconstruction: Using the point cloud data and the EIT measurements obtained from real experiments to perform reconstruction with the VD2T model.
\end{enumerate}
This procedure results in a geometrical reconstruction of the deformed body along with the tactile map.

\subsection{Architecture of VD2T}
We propose the Voltage and Deformation to Tactile (VD2T) model to address deformable tactile reconstruction using EIT. Existing EIT-based tactile reconstruction methods, which use measurements taken before deformation and without touch as references, tend to result in significant errors in cases with deformable surfaces. It is often impractical to acquire suitable references during deformations. Moreover, incorporating deformation directly into the reconstruction process without relying on a time-consuming forward modelling process remains challenging. VD2T is designed to overcome these limitations by directly using the EIT measurements obtained before deformation and without touch as the reference, along with the point cloud data depicting the sensor morphology, to reconstruct the tactile map on deformed surfaces.

Fig. \ref{Fig3} also shows the architecture of the VD2T model. This consists of two channels that effectively extract and fuse features from EIT measurements and point cloud data providing a spatial cartesian estimation of the deformed target object. The first channel employs a three-layer Multilayer Perceptron (MLP) to extract key features of EIT measurements. The MLP is essential to capture global signal characteristics, while the subsequent three-layer one-dimensional Convolutional Neural Network (CNN) focuses on extracting fine-grained local patterns. This hierarchical design ensures that both global and local information are effectively utilized, enhancing the tactile reconstruction accuracy. The second channel extracts effective features of three-dimensional spatial structure from point cloud data through the two-layer two-dimensional CNN. After feature extraction, the two features are fused through a deep fully connected network that includes two-layer linear transformations, and the output layer uses the Sigmoid activation function to map the results to the final prediction space. To accelerate the training process and improve model performance, VD2T adopts a Batch Normalization strategy \cite{Ioffe2015}. In addition, by introducing the dropout strategy, the model effectively alleviates the overfitting problem and enhances its generalization ability on unseen data.

To train the VD2T model, we utilize the Binary Cross-Entropy Loss function, 
which is defined as:
\begin{equation}
L(y, \hat{y})=-\frac{1}{N} \sum_{i=1}^N\left[y_i \cdot \log \left(\hat{y}_i\right)+\left(1-y_i\right) \cdot \log \left(1-\hat{y}_i\right)\right]
\label{eq5}
\end{equation}
where ${y}_i$ represents the true label of sample $i$, $\hat{y}_i$ is the model's predicted probability for sample $i$, and $N$ is the total number of samples in the batch. 


\begin{figure}[t]
\centering
\includegraphics[scale=0.7]{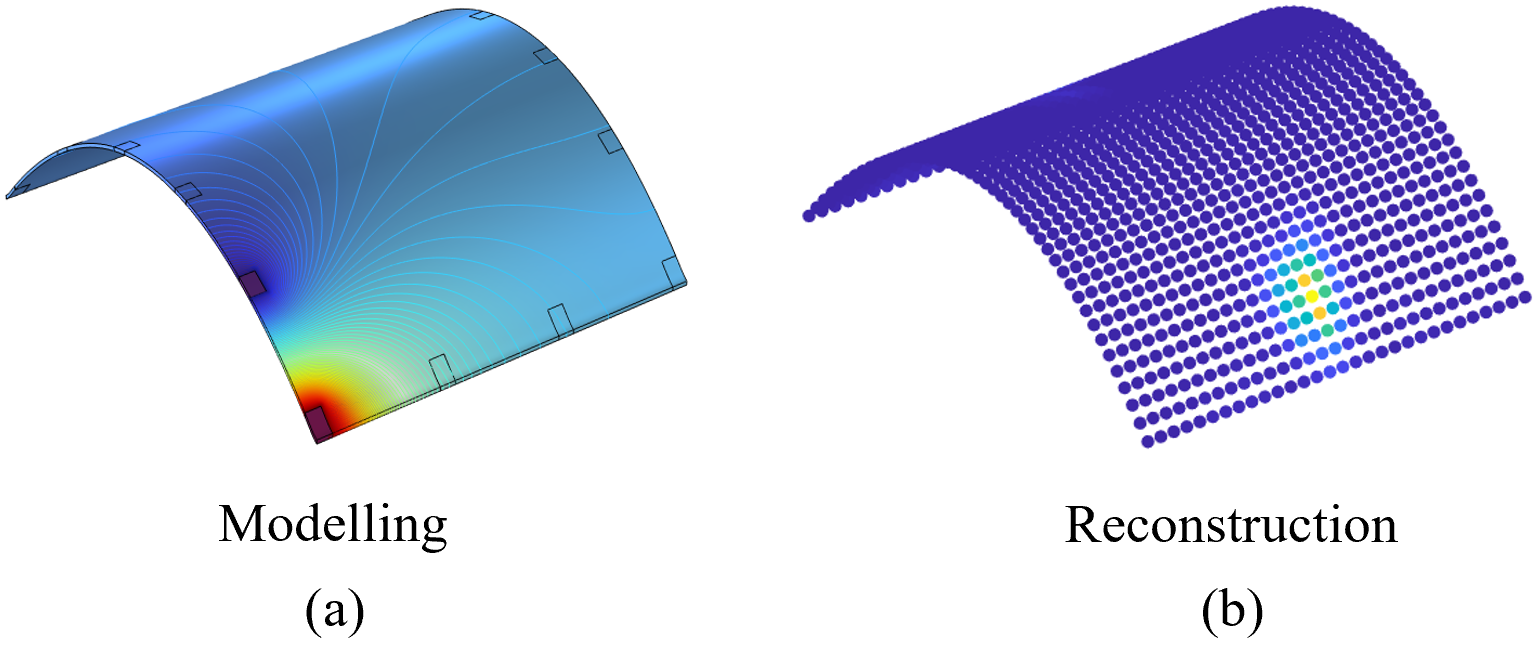}
\caption{Tactile reconstruction on deformed surface. (a) Initial finite element model. (b) Tactile reconstruction based on the traditional method.}
\label{Fig4}
\end{figure}

\subsection{Traditional EIT algorithms under Deformed Surfaces}
Few existing EIT-based tactile sensing approaches have explored cases involving sensor deformations. To compare the proposed VD2T with existing EIT algorithms in deformable tactile sensing, it is necessary to modify existing methods to incorporate deformation information during forward EIT-reconstruction modelling. To achieve this, the point cloud data from the 3D scan is employed to establish the initial finite element model and define the model's geometry. Building upon the initial model, we introduce additional components such as electrodes and assign appropriate material properties to emulate real-world conditions. This process results in a comprehensive model that facilitates subsequent simulation calculations (see Fig. 4a). Using this refined model enables the computation of the updated sensitivity (or Jacobian) matrix. The sensitivity matrix, along with the real-world EIT measurements, allows the reconstruction of the touch areas with existing algorithms (see Fig. 4b). Note that here we first employ EIT measurements after deformation but without touch as the reference and then analyze the impact in the later section. 

Mesh configuration is a key aspect of our numerical simulations. To avoid inverse crime, we employ a finer mesh for the model and a coarser mesh for point cloud extraction in sensitivity matrix calculation  \cite{Lionheart2004}. In the example (see Fig. \ref{Fig4}), the model is depicted with a point cloud with 8671 points, while the number of points used for sensitivity matrix calculation is 1350. 
\begin{figure}[t]
\centerline{\includegraphics[scale=0.15]{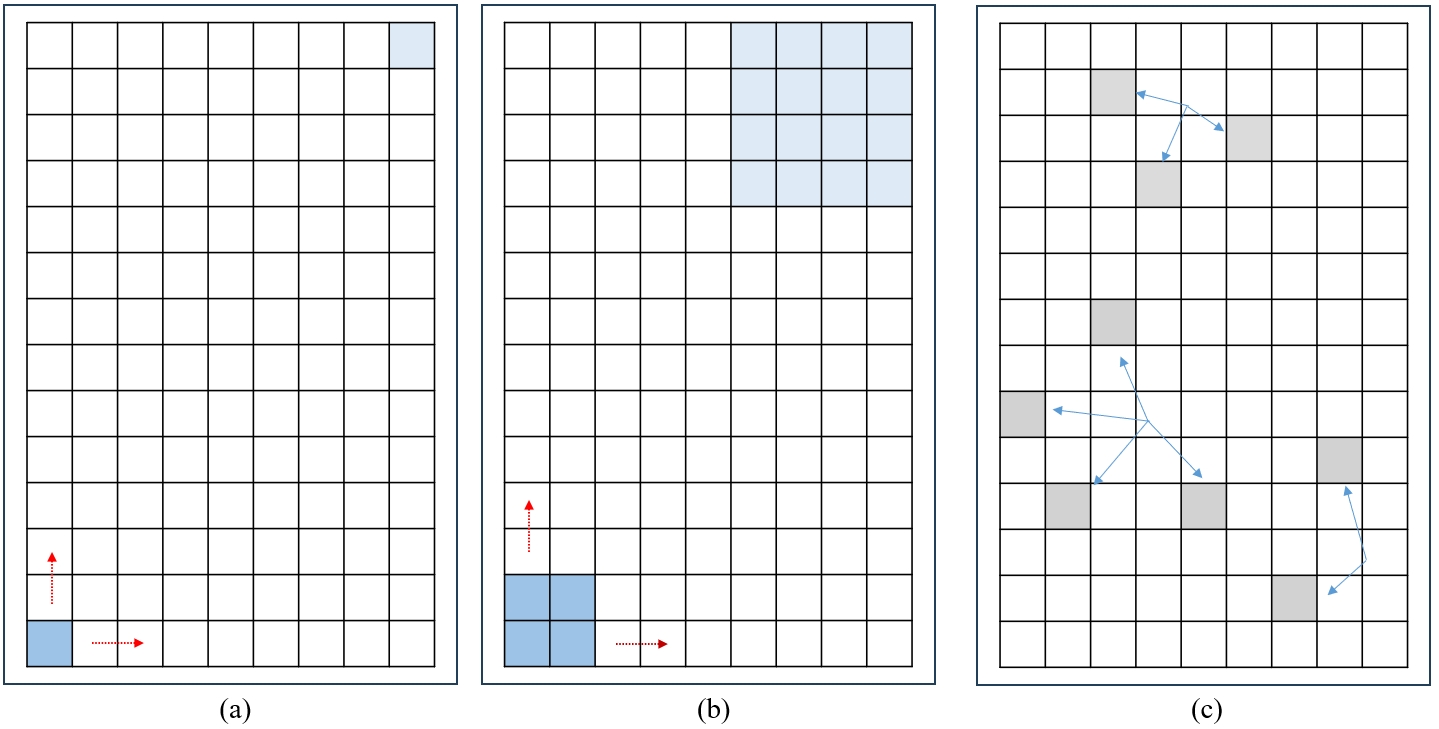}}
\caption{Three types of touch patterns in the dataset. (a) One unit: Each unit is touched individually and sequentially. (b) Square-shaped area: Touch is applied to square areas ranging from 2 $\times$ 2 to 9 $\times$ 9 units sequentially. (c) Random 2 to 4 units.}
\label{Fig5}
\end{figure}

\section{Numerical Simulation}
\subsection{Dataset and Network Training}
We generate a deformable tactile sensing dataset based on multiphysics simulation for network training. As shown in Fig. \ref{Fig5}, we divide the sensing region with a size of 100 $\times$ 150 mm${^2}$ into 126 units (9 $\times$ 14) and set the background conductivity to 1 S/m and the touch area conductivity to 50 S/m to simulate the touch effect. We include three types of touch patterns in the dataset, similar to  \cite{Park2021}. The first touch pattern is to iterate over each unit and designate it as a touch area, thereby overlaying the impacts of each unit under analysis. The second pattern is square areas of different sizes, from 2 $\times$ 2 to 9 $\times$ 9, to gather information about the properties of cluster and surface distributions. The third pattern is randomly selecting 2 to 4 units for touch, the interdependence of random units is studied through the perturbation of random units. To capture the effects of mechanical deformation on the tactile sensor's response,  all three touch patterns are applied to the sensor under different bending deformation states. The deformation simulated a bending process where the EIT e-skin is gradually curved to resemble a cylindrical shape. This was achieved by altering the specified displacement parameters in multiphysics simulations, systematically bending the sensor from a flat state to various degrees of curvature, and mimicking realistic operational conditions. 
Finally, we generated a dataset of 102,590 samples, each mapping 104 voltage measurements to 1350 conductivity values. Data processing and simulations were performed using MATLAB and COMSOL Multiphysics.
\begin{figure}[t]
\centerline{\includegraphics[width=\columnwidth]{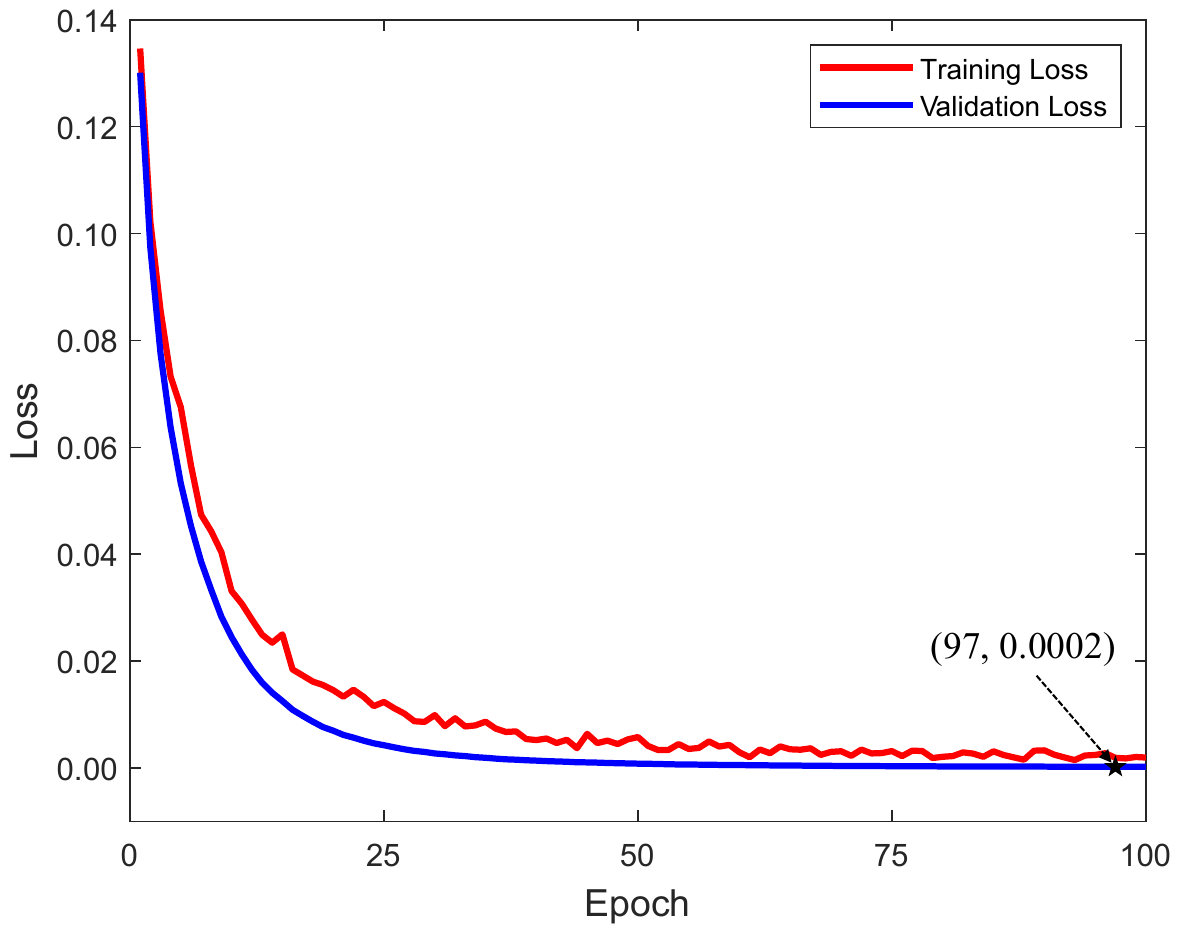}}
\caption{Training and validation loss of the VD2T model.}
\label{fig-losscurve}
\end{figure}

\begin{figure*}[t]
\centering
\includegraphics[scale=0.72]{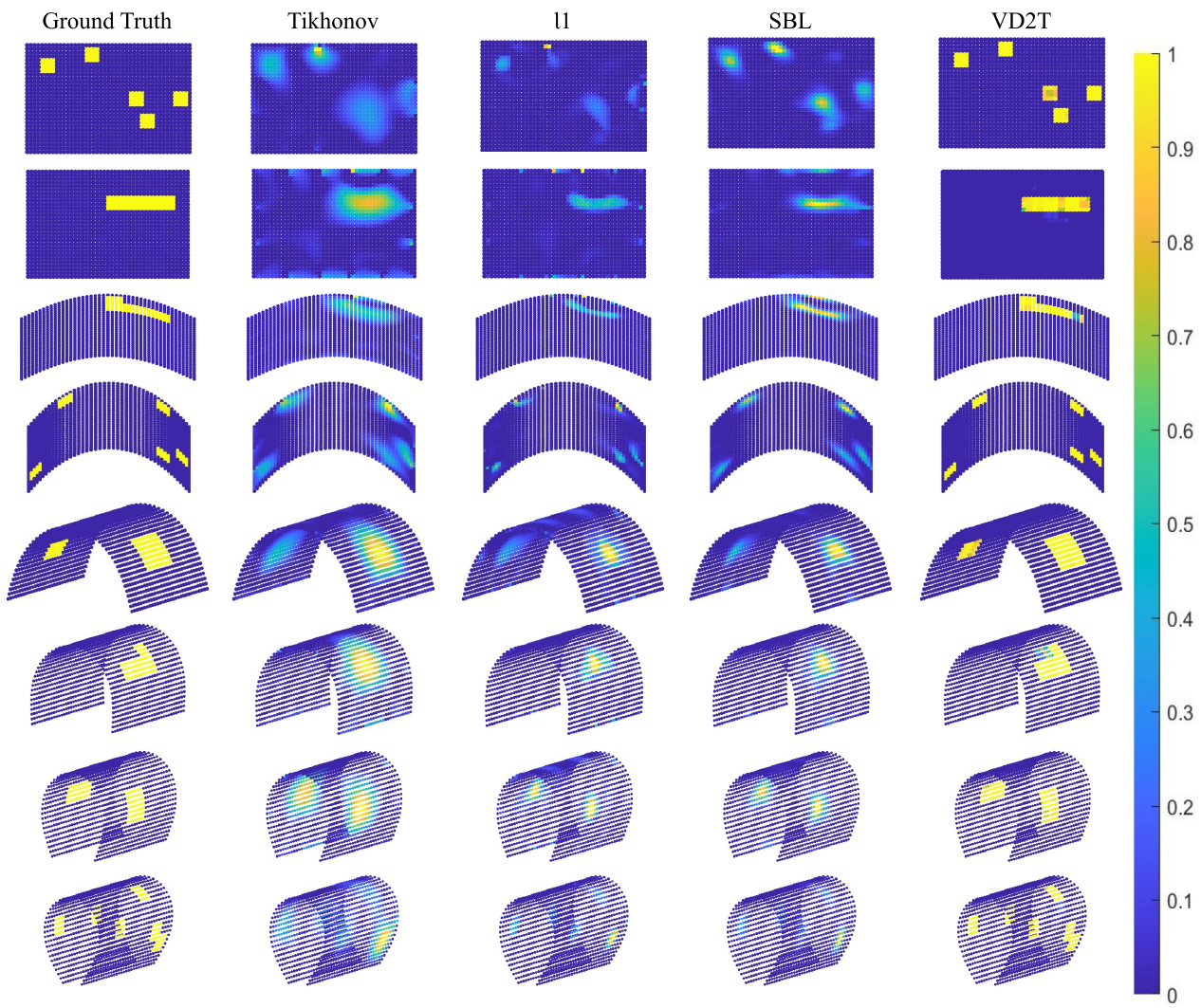}
\caption{Tactile reconstruction on deformed surfaces based on simulation data (SNR = 50 dB).}
\label{Fig7}
\end{figure*}
The VD2T model was trained using simulated point cloud data and the relative change in voltage measurements as inputs and the corresponding conductivity distribution induced by touch served as the target output. We employed the Adam optimizer \cite{Kingma2014} with a learning rate of 0.0001 for training 100 epochs, with a batch size of 512. Of the total dataset, the training and validation datasets were split with a 90/10 ratio, respectively. Notably, the VD2T model was trained exclusively on simulated datasets due to the challenge of obtaining ground truth in real-world experiments. Experimental setups often lack access to tactile ground truth due to hardware limitations or the complexity of direct measurement.  Simulations, on the other hand, can provide ground truth labels while capturing the physical principles of conductivity and voltage relationships underlying EIT.  To further improve the generalization ability, additive white Gaussian noise with a Signal-to-Noise Ratio (SNR) of 50dB was added to the training data for data augmentation, emulating real-world disturbances commonly encountered in experimental hardware.  This noise injection ensures the model's robustness and adaptability to experimental scenarios. For testing, we used different test data (Section V.C). Fig. \ref{fig-losscurve} shows the training and validation loss. The network underwent training for 100 epochs and achieved the minimum validation loss of 0.0002 at the 97th epoch, which was preserved as the optimal model. All training and simulations were conducted on a computer equipped with an Intel i9-13900 CPU and an RTX 4080 GPU.

\subsection{Parameters of Comparison Algorithms}
We compared VD2T with established regularization techniques, including classical Tikhonov (Tikh) regularization \cite{Vauhkonen1998}, $l1$ regularization \cite{Tehrani2012}, and Sparse Bayesian Learning (SBL) \cite{Liu2019}. 

Parameters of these algorithms were identified using trial and error to ensure a fair comparison. The regularization factor of Tikh was set to 0.001. The $l1$ regularization employed a regularization factor of 0.001 with a maximum iteration number of 400. The SBL was configured with a maximum iteration number of 5, cluster size of 4, tolerance of 0.0001 and pattern coupling factor of 0.3. For comparison algorithms, the reference EIT measurement for calculating the relative change was measured on the deformed surface but before the touch was implemented. For VD2T, the reference voltage was measured before any deformation and touch occurred.

\subsection{Results and Discussion}
To evaluate the performance of given approaches for deformable tactile sensing, we employed testing samples with entirely different touch shapes and numbers from the training set. Eight touch phantoms were introduced, ranging from flat to highly curved surfaces (see Fig. 7). In Phantoms 1, we applied five discrete touch points, each measuring 10 $\times$ 10 mm$^{2}$. For Phantom 2, we created the rectangular touch areas measuring 10 $\times$ 60 mm$^{2}$.  In Phantom 3, we designed the L-shaped touch area of six 10 $\times$ 10 mm$^{2}$ touch points. For Phantom 4 we applied five discrete touch points, each measuring 10 $\times$ 10 mm$^{2}$. For Phantom 5, we created two rectangular touch areas measuring 20 $\times$ 10 mm$^{2}$ and 20 $\times$ 30 mm$^{2}$, respectively. In Phantom 6, we designed an L-shaped touch area composed of five 10 $\times$ 10 mm$^{2}$ touch points. Phantom 7 featured a 30 $\times$ 10 mm$^{2}$ rectangle alongside a 20 $\times$ 20 mm$^{2}$ square, and Phantom 8 included six discrete touch points, each 10 $\times$ 10 mm$^{2}$. These phantoms were specifically chosen to evaluate the generalization ability of VD2T, as they differ significantly from those used in the training dataset in geometries and the number of touch areas. We can observe that the VD2T can reconstruct the touch positions, sizes, and shapes well across all eight phantoms. This demonstrates the robustness of the VD2T model in handling unseen touch patterns and complex geometries, even those absent from the training dataset.

We employ the Correlation Coefficient (CC), average peak SNR (PSNR) and Relative Image Error (RIE) to assess the tactile reconstruction performance \cite{Dong2024}. These quantitative metrics (see Table \ref{table-qm}) show higher accuracy and superior performance of VD2T compared with other given algorithms, which is evidenced by the high CC ranging from 0.9660 to 0.9999. Similarly, the PSNR varying between 28.7221 and 55.5264 and the RIE spanning from 0.0107 to 0.0805 further validated VD2T's superior performance in reconstructing fine details of tactile interactions on deformed surfaces without requiring a reference measurement after deformation.

\begin{table}[t]
\centering
\caption{Quantitative metrics of four tactile reconstruction algorithms.}
\label{table-qm}
\setlength{\tabcolsep}{10pt}
\begin{tabular}{|cc|c|c|c|c|}
\hline
\multicolumn{2}{|c|}{Phantom}                   & Tikhonov    & $l1$      & SBL     & VD2T    \\ \hline
\multicolumn{1}{|c|}{\multirow{3}{*}{1}} & CC   & 0.5967  & 0.5755  & 0.6877  & \textbf{0.9976}  \\ \cline{2-6} 
\multicolumn{1}{|c|}{}                   & PSNR & 18.1990 & 17.2960 & 18.9786 & \textbf{38.9502} \\ \cline{2-6} 
\multicolumn{1}{|c|}{}                   & RIE  & 0.7816  & 0.8672  & 0.7145  & \textbf{0.0737}   \\ \hline
\multicolumn{1}{|c|}{\multirow{3}{*}{2}} & CC   & 0.7198  & 0.6080  & 0.7110  & \textbf{0.9762}  \\ \cline{2-6} 
\multicolumn{1}{|c|}{}                   & PSNR & 19.4275 & 17.8148 & 19.3868 & \textbf{29.6559} \\ \cline{2-6} 
\multicolumn{1}{|c|}{}                   & RIE  & 0.6961  & 0.8381  & 0.6994  & \textbf{0.2282}  \\ \hline
\multicolumn{1}{|c|}{\multirow{3}{*}{3}} & CC   & 0.6387  & 0.6700  & 0.7408  & \textbf{0.9957}  \\ \cline{2-6} 
\multicolumn{1}{|c|}{}                   & PSNR & 18.3340 & 18.4495 & 19.4744 & \textbf{36.2172} \\ \cline{2-6} 
\multicolumn{1}{|c|}{}                   & RIE  & 0.7895  & 0.7791  & 0.6924  & \textbf{0.1061}  \\ \hline
\multicolumn{1}{|c|}{\multirow{3}{*}{4}} & CC   & 0.6565  & 0.6484  & 0.8132  & \textbf{0.9999}  \\ \cline{2-6} 
\multicolumn{1}{|c|}{}                   & PSNR & 18.7610 & 18.2117 & 19.6773 & \textbf{55.5264} \\ \cline{2-6} 
\multicolumn{1}{|c|}{}                   & RIE  & 0.7326  & 0.7804  & 0.6593  & \textbf{0.0107}  \\ \hline
\multicolumn{1}{|c|}{\multirow{3}{*}{5}} & CC   & 0.7641  & 0.7118  & 0.8020  & \textbf{0.9987}  \\ \cline{2-6} 
\multicolumn{1}{|c|}{}                   & PSNR & 19.2747 & 18.0408 & 18.6075 & \textbf{40.5856} \\ \cline{2-6} 
\multicolumn{1}{|c|}{}                   & RIE  & 0.6239  & 0.7191  & 0.6737  & \textbf{0.0549}  \\ \hline
\multicolumn{1}{|c|}{\multirow{3}{*}{6}} & CC   & 0.6659  & 0.7010  & 0.7085  & \textbf{0.9660}  \\ \cline{2-6} 
\multicolumn{1}{|c|}{}                   & PSNR & 18.6874 & 19.9576 & 20.0314 & \textbf{28.7221} \\ \cline{2-6} 
\multicolumn{1}{|c|}{}                   & RIE  & 0.8260  & 0.7137  & 0.7076  & \textbf{0.2526}  \\ \hline
\multicolumn{1}{|c|}{\multirow{3}{*}{7}} & CC   & 0.7036  & 0.7395  & 0.8152  & \textbf{0.9974}  \\ \cline{2-6} 
\multicolumn{1}{|c|}{}                   & PSNR & 17.9896 & 19.1457 & 19.7226 & \textbf{37.8094} \\ \cline{2-6} 
\multicolumn{1}{|c|}{}                   & RIE  & 0.7591  & 0.6645  & 0.6218  & \textbf{0.0805}  \\ \hline
\multicolumn{1}{|c|}{\multirow{3}{*}{8}} & CC   & 0.5551  & 0.6665  & 0.6549  & \textbf{0.9981}  \\ \cline{2-6} 
\multicolumn{1}{|c|}{}                   & PSNR & 17.1949 & 17.2887 & 17.1671 & \textbf{39.6031} \\ \cline{2-6} 
\multicolumn{1}{|c|}{}                   & RIE  & 0.8051  & 0.7965  & 0.8077  & \textbf{0.0618}  \\ \hline
\end{tabular}
\vspace{0.2mm}
\begin{minipage}{\columnwidth}
\raggedright
\footnotesize{The best results are highlighted in bold.}
\end{minipage}
\end{table}
\begin{figure*}[t]
\centering
\includegraphics[scale=0.70]{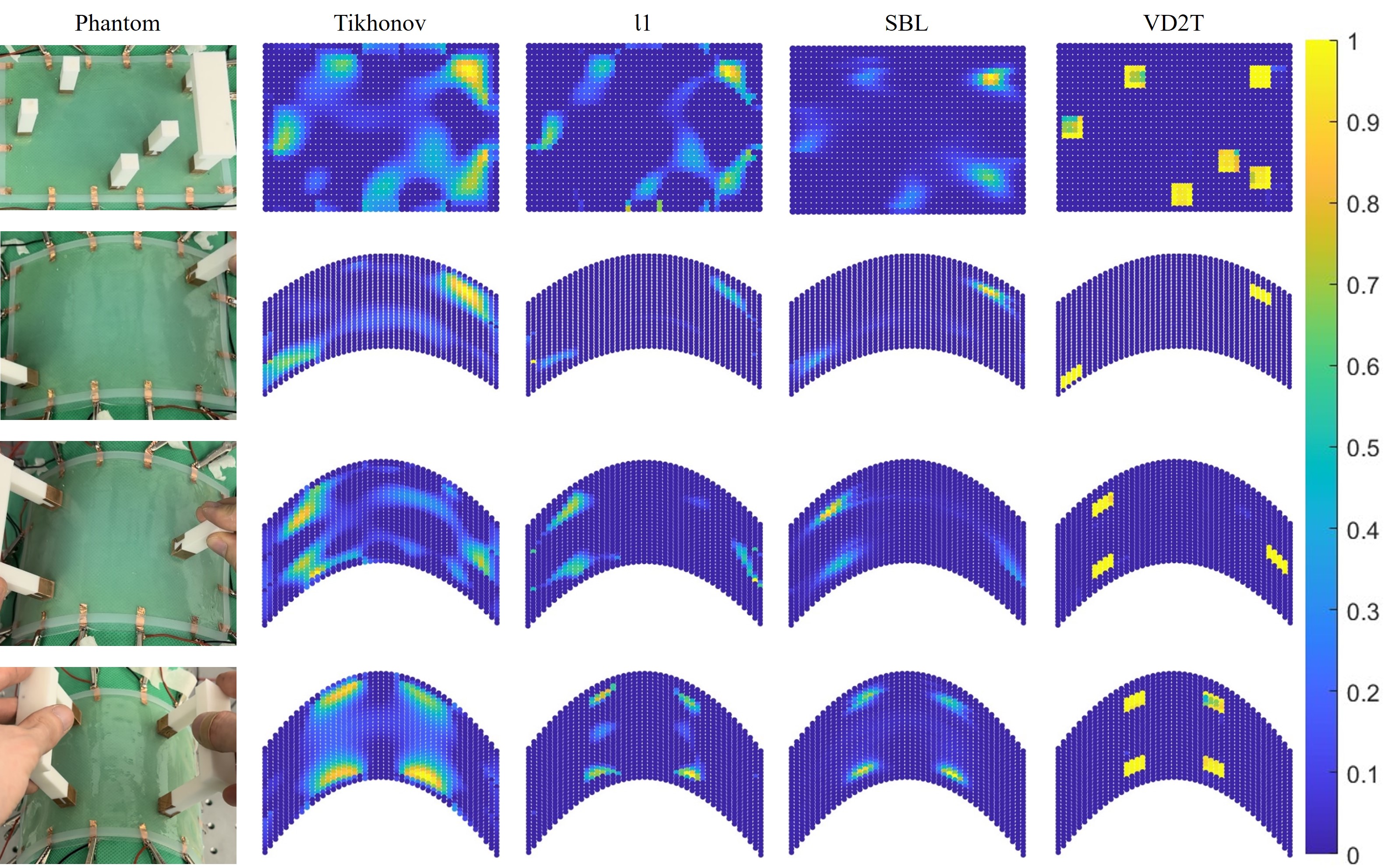}
\caption{Deformable tactile reconstruction results based on experiment data.}
\label{Fig8}
\end{figure*}
\section{Real-world Experiments}
\subsection{Experimental Settings}
We conducted real-world experiments to assess the practical performance of the proposed approach. We employed the previously described hydrogel-based tactile e-skin and placed it on a flexible foam pad to generate varying degrees of bending. This setup allows for systematic control of the curvature of the pad to induce specific deformations of the e-skin. Regularization techniques (e.g., Tikh, $l1$, SBL) use reference voltages measured post-deformation to compensate for deformation effects, which can improve performance under higher degrees of bending. However, VD2T requires only a single pre-deformation reference, demonstrating its robustness across different bending scenarios without additional reference adjustments. Finally, we collected experimental data under four distinct deformation scenarios. A multi-frequency EIT system was used for data collection and adjacent-excitation and adjacent measurement protocol was adopted \cite{Yang2017}, where the current frequency is 20 kHz. Each scenario involved a different phantom (deformed shape) with a varying number of designated touch points (i.e., 6, 2, 3, and 4). For consistency and ease of quantification, all touch points across the objects were standardized as combinations of 10$\times$10 mm$^{2}$ squares.

Algorithm parameters were determined through a trial-and-error process for optimized performance, which is listed in Table \ref{tab2}. Similar to the simulation setup, the selection of reference voltage differs between these regularization methods and VD2T. Traditional methods utilized the voltage measured post-deformation as the reference to minimize the inference caused by deformation. In contrast, VD2T is capable of using the pre-deformation readout as the reference. This means that VD2T requires only a single acquisition of the reference voltage before any deformation or touch. Therefore, VD2T can effectively address the limitation of requiring varying reference voltage during the deformation.
\begin{table}[t]
\centering
\caption{PARAMETERS OF COMPARISON ALGORITHMS}
\setlength{\tabcolsep}{5.5pt}
\begin{tabular}{|c|c|cc|cccc|}
\hline
\multirow{2}{*}{Phan.} & Tikh    & \multicolumn{2}{c|}{$l1$}                                           & \multicolumn{4}{c|}{SBL}                                                                                                                                \\ \cline{2-8} 
                         & factor 1 & \multicolumn{1}{c|}{factor 1}                & num                  & \multicolumn{1}{c|}{num}                & \multicolumn{1}{c|}{size}               & \multicolumn{1}{c|}{tolerance}               & factor 2              \\ \hline
1                        & 0.001   & \multicolumn{1}{c|}{0.3}                   & \multirow{3}{*}{300} & \multicolumn{1}{c|}{\multirow{4}{*}{5}} & \multicolumn{1}{c|}{\multirow{4}{*}{4}} & \multicolumn{1}{c|}{\multirow{4}{*}{0.0001}} & \multirow{4}{*}{0.3} \\ \cline{1-3}
2                        & 0.01    & \multicolumn{1}{c|}{\multirow{3}{*}{0.45}} &                      & \multicolumn{1}{c|}{}                   & \multicolumn{1}{c|}{}                   & \multicolumn{1}{c|}{}                        &                      \\ \cline{1-2}
3                        & 0.001   & \multicolumn{1}{c|}{}                      &                      & \multicolumn{1}{c|}{}                   & \multicolumn{1}{c|}{}                   & \multicolumn{1}{c|}{}                        &                      \\ \cline{1-2} \cline{4-4}
4                        & 0.01    & \multicolumn{1}{c|}{}                      & 100                  & \multicolumn{1}{c|}{}                   & \multicolumn{1}{c|}{}                   & \multicolumn{1}{c|}{}                        &                      \\ \hline

\end{tabular}
\label{tab2}
\vspace{0.2mm}
\begin{minipage}{\columnwidth}
\footnotesize Factor 1: regularization factor; num: maximum iteration number; size: cluster size; factor 2: pattern coupling factor.
\end{minipage}
\end{table}
\begin{figure}[t]
\centerline{\includegraphics[scale=0.7]{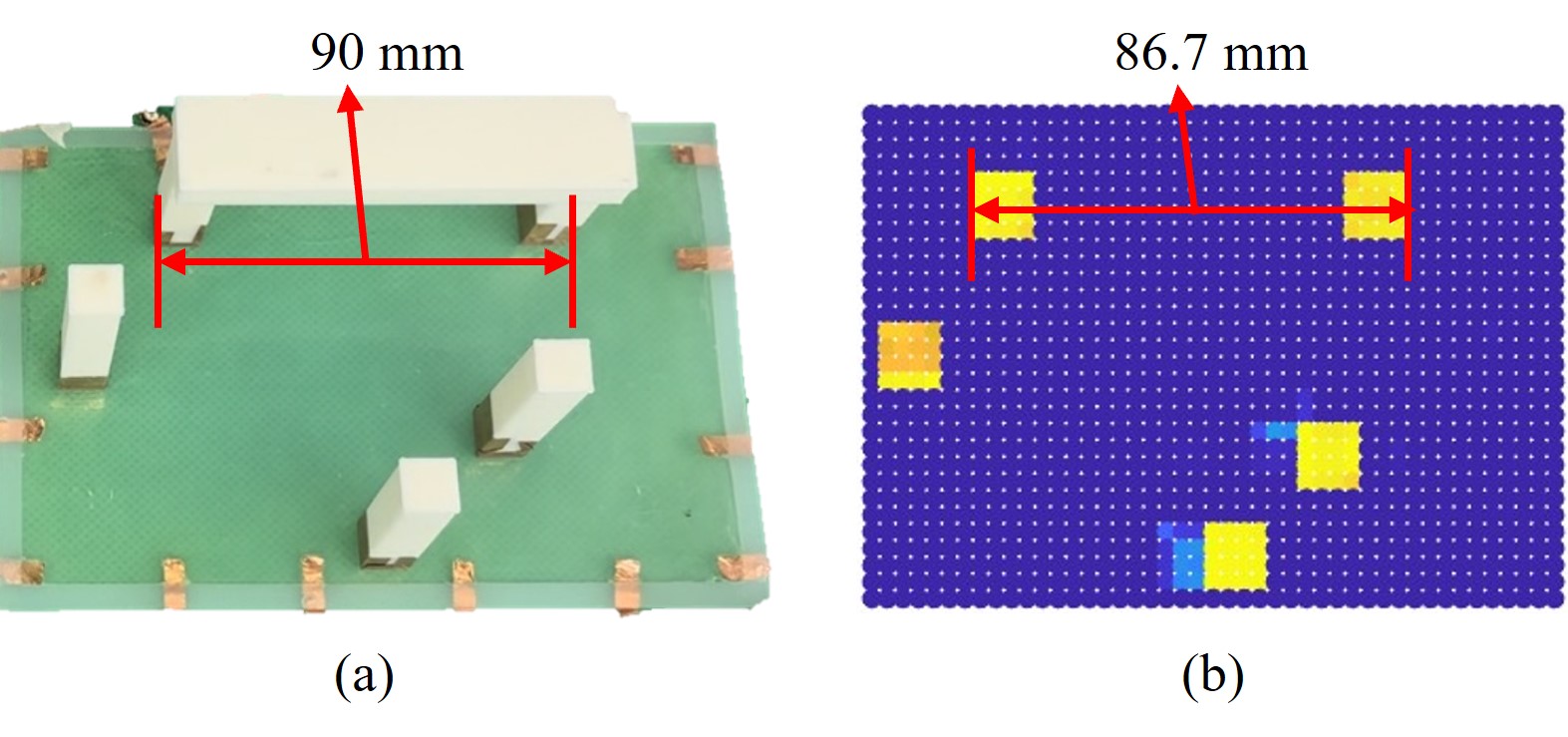}}
\caption{Quantitative evaluation using DE. (a) The distance between
the boundaries of two touch areas in the real sensor. (b) The distance between
the boundaries of two touch areas in the reconstruction. }
\label{Fig9}
\end{figure}

\subsection{Quantitative Metrics}
Unlike simulations where ground truth data is readily available, real-world experiments require alternative evaluation metrics. Here, we employ the Distance Error (DE) to assess the quality of the reconstructed touch areas \cite{shooter2005use}:
\begin{equation}
\mathrm{DE}=\frac{\left|D-D_r\right|}{D_r}
\label{eq8}
\end{equation}
where $D$ is the distance between the boundaries of two touch areas within the reconstruction result, and $D_r$ is the true distance between the boundaries of two touch areas.

\subsection{Results and Discussion}
Fig. \ref{Fig8} shows the tactile reconstruction results from the given methods. While Tikh, \textit{l}1 regularization, and SBL algorithms partially reconstruct the number and position of the touch areas, they exhibit significant noise and fail to capture the shape effectively. In contrast, VD2T can successfully reconstruct the number, position and shape of the touch areas and effectively suppress the artefacts, achieving high-quality reconstructions.

To quantify the performance of VD2T in real-world scenarios (where ground truth is unavailable), we set up five touch areas on the hydrogel e-skin (see Fig. \ref{Fig9}) and evaluate DE. The distance between the boundaries of the two touch areas is fixed at 90 mm. VD2T can successfully recover these areas, with a measured distance of 86.7 mm between the two closest. This translates to a DE of 3.37\%, suggesting the good generalization ability of VD2T to experimental data.
\begin{figure*}[t]
\centerline{\includegraphics[width=\textwidth]{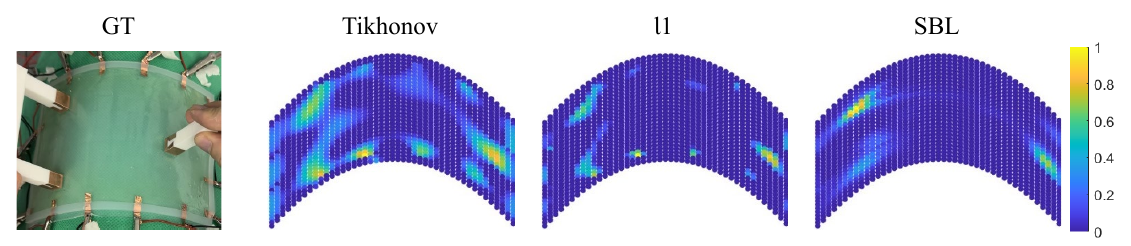}}
\caption{Tactile reconstruction using the EIT measurements before deformation and without touch as reference.}
\label{Fig10}
\end{figure*}

As aforementioned, for time-difference EIT-based tactile sensing, it is impractical to obtain a proper reference during deformations. The use of the undeformed reference can induce non-negligible tactile reconstruction errors due to the inference from deformations. Here we further demonstrate the impact of adopting undeformed reference on tactile reconstructions for existing algorithms. It is noteworthy that VD2T takes into account this effect and employs the reference taken under undeformed conditions and mitigates its influence on tactile reconstruction through the learning process, which cannot be achieved by existing approaches. Fig. \ref{Fig10} illustrates the reconstruction results from comparison algorithms for Phantom 3 using EIT measurements before deformation and without touch as the reference. The reconstructions using the undeformed reference voltage show more artefacts and degraded image quality. This performance degradation is attributed to the mismatch between the reference voltage and the actual deformed state of the e-skin. This result further confirms the superiority of VD2T, which only requires a pre-deformation reference, thereby being more practical in real-world scenarios.

\section*{Conclusion}
This paper proposes a tactile reconstruction approach that tracks and effectively compensates for surface deformations during tactile sensing. We first capture the deformations of the e-skin through a high-precision 3D scan and then perform tactile reconstruction using the VD2T model, which is specifically designed to process and fuse EIT data and deformation information. This model addresses the limitation of existing EIT-based tactile reconstruction algorithms, which makes it challenging to obtain proper reference during deformation and incorporate deformed states in the estimation. VD2T directly uses the EIT measurements before deformation and without touch as the reference and in the meanwhile it leverages point cloud information to reconstruct the tactile map on deformed surfaces. Results from simulations and experiments confirm the method's capability for tactile sensing on deformable surfaces. Future research will focus on developing integrated flexible sensors that can directly capture deformations in real-time, thereby eliminating the need for external scanners to acquire deformations, and exploring the applications in soft and highly deformable robotic systems. In addition, we state that the proposed e-skin is primarily designed for contact localization and contact shape reconstruction on large-area and deformed surfaces. While normal or shear stress is not differentiated, our future work will investigate stress-sensing capabilities by developing multi-layered sensor designs and expanding the application scopes.

\bibliographystyle{IEEEtranTIE}
\bibliography{References}\ 

\begin{thebibliography}{10}
\providecommand{\url}[1]{#1}
\csname url@samestyle\endcsname
\providecommand{\newblock}{\relax}
\providecommand{\bibinfo}[2]{#2}
\providecommand{\BIBentrySTDinterwordspacing}{\spaceskip=0pt\relax}
\providecommand{\BIBentryALTinterwordstretchfactor}{4}
\providecommand{\BIBentryALTinterwordspacing}{\spaceskip=\fontdimen2\font plus
\BIBentryALTinterwordstretchfactor\fontdimen3\font minus \fontdimen4\font\relax}
\providecommand{\BIBforeignlanguage}[2]{{%
\expandafter\ifx\csname l@#1\endcsname\relax
\typeout{** WARNING: IEEEtran.bst: No hyphenation pattern has been}%
\typeout{** loaded for the language `#1'. Using the pattern for}%
\typeout{** the default language instead.}%
\else
\language=\csname l@#1\endcsname
\fi
#2}}
\providecommand{\BIBdecl}{\relax}
\BIBdecl

\bibitem{Loomis1986}
J.~M. Loomis and S.~J. Lederman, ``Tactual perception,'' \emph{Handbook of perception and human performances}, vol.~2, no.~2, p.~2, 1986.

\bibitem{Booth2018}
J.~W. Booth, D.~Shah, J.~C. Case, E.~L. White, M.~C. Yuen, O.~Cyr-Choiniere, and R.~Kramer-Bottiglio, ``Omniskins: Robotic skins that turn inanimate objects into multifunctional robots,'' \emph{Science Robotics}, vol.~3, no.~22, p. eaat1853, 2018.

\bibitem{Yang2023}
M.~J. Yang, K.~Park, W.~D. Kim, and J.~Kim, ``Robotic skin mimicking human skin layer and pacinian corpuscle for social interaction,'' \emph{IEEE/ASME Transactions on Mechatronics}, 2023.

\bibitem{Cheng2019}
G.~Cheng, E.~Dean-Leon, F.~Bergner, J.~R.~G. Olvera, Q.~Leboutet, and P.~Mittendorfer, ``A comprehensive realization of robot skin: Sensors, sensing, control, and applications,'' \emph{Proc. IEEE}, vol. 107, no.~10, pp. 2034--2051, 2019.

\bibitem{SilveraTawil2015}
D.~Silvera-Tawil, D.~Rye, and M.~Velonaki, ``Artificial skin and tactile sensing for socially interactive robots: A review,'' \emph{Robotics and Autonomous Systems}, vol.~63, pp. 230--243, 2015.

\bibitem{Sundaram2019}
S.~Sundaram, P.~Kellnhofer, Y.~Li, J.-Y. Zhu, A.~Torralba, and W.~Matusik, ``Learning the signatures of the human grasp using a scalable tactile glove,'' \emph{Nature}, vol. 569, no. 7758, pp. 698--702, 2019.

\bibitem{Dahiya2013_2}
R.~S. Dahiya and M.~Valle, \emph{Robotic tactile sensing: technologies and system}, vol.~1.\hskip 1em plus 0.5em minus 0.4em\relax Springer, 2013.

\bibitem{Boutry2018}
C.~M. Boutry, M.~Negre, M.~Jorda, O.~Vardoulis, A.~Chortos, O.~Khatib, and Z.~Bao, ``A hierarchically patterned, bioinspired e-skin able to detect the direction of applied pressure for robotics,'' \emph{Science Robotics}, vol.~3, no.~24, p. eaau6914, 2018.

\bibitem{Won2019}
S.~M. Won, H.~Wang, B.~H. Kim, K.~Lee, H.~Jang, K.~Kwon, M.~Han, K.~E. Crawford, H.~Li, Y.~Lee \emph{et~al.}, ``Multimodal sensing with a three-dimensional piezoresistive structure,'' \emph{ACS nano}, vol.~13, no.~10, pp. 10\,972--10\,979, 2019.

\bibitem{Park2021}
H.~Park, K.~Park, S.~Mo, and J.~Kim, ``Deep neural network based electrical impedance tomographic sensing methodology for large-area robotic tactile sensing,'' \emph{IEEE Transactions on Robotics}, vol.~37, no.~5, pp. 1570--1583, 2021.

\bibitem{Tawil2011}
D.~S. Tawil, D.~Rye, and M.~Velonaki, ``Improved image reconstruction for an eit-based sensitive skin with multiple internal electrodes,'' \emph{IEEE Transactions on Robotics}, vol.~27, no.~3, pp. 425--435, 2011.

\bibitem{Kato2007}
Y.~Kato, T.~Mukai, T.~Hayakawa, and T.~Shibata, ``Tactile sensor without wire and sensing element in the tactile region based on eit method,'' in \emph{SENSORS, 2007 IEEE}, pp. 792--795.\hskip 1em plus 0.5em minus 0.4em\relax IEEE, 2007.

\bibitem{Nagakubo2007}
A.~Nagakubo, H.~Alirezaei, and Y.~Kuniyoshi, ``A deformable and deformation sensitive tactile distribution sensor,'' in \emph{2007 IEEE International Conference on Robotics and Biomimetics (ROBIO)}, pp. 1301--1308.\hskip 1em plus 0.5em minus 0.4em\relax IEEE, 2007.

\bibitem{Liu2015}
D.~Liu, V.~Kolehmainen, S.~Siltanen, A.-M. Laukkanen, and A.~Sepp{\"a}nen, ``Nonlinear difference imaging approach to three-dimensional electrical impedance tomography in the presence of geometric modeling errors,'' \emph{IEEE Transactions on Biomedical Engineering}, vol.~63, no.~9, pp. 1956--1965, 2015.

\bibitem{Alirezaei2009}
H.~Alirezaei, A.~Nagakubo, and Y.~Kuniyoshi, ``A tactile distribution sensor which enables stable measurement under high and dynamic stretch,'' in \emph{2009 IEEE Symposium on 3D User Interfaces}, pp. 87--93.\hskip 1em plus 0.5em minus 0.4em\relax IEEE, 2009.

\bibitem{Li2024}
Y.~Li, G.~Matsumura, Y.~Xuan, S.~Honda, and K.~Takei, ``Stretchable electronic skin using laser-induced graphene and liquid metal with an action recognition system powered by machine learning,'' \emph{Advanced Functional Materials}, p. 2313824, 2024.

\bibitem{Kuen2009}
J.~Kuen, E.~J. Woo, and J.~K. Seo, ``Multi-frequency time-difference complex conductivity imaging of canine and human lungs using the khu mark1 eit system,'' \emph{Physiological measurement}, vol.~30, no.~6, p. S149, 2009.

\bibitem{Zhu2020}
Z.~Zhu, H.~S. Park, and M.~C. McAlpine, ``3d printed deformable sensors,'' \emph{Science advances}, vol.~6, no.~25, p. eaba5575, 2020.

\bibitem{Park2022}
K.~Park, H.~Yuk, M.~Yang, J.~Cho, H.~Lee, and J.~Kim, ``A biomimetic elastomeric robot skin using electrical impedance and acoustic tomography for tactile sensing,'' \emph{Science Robotics}, vol.~7, no.~67, p. eabm7187, 2022.

\bibitem{Keplinger2013}
C.~Keplinger, J.-Y. Sun, C.~C. Foo, P.~Rothemund, G.~M. Whitesides, and Z.~Suo, ``Stretchable, transparent, ionic conductors,'' \emph{Science}, vol. 341, no. 6149, pp. 984--987, 2013.

\bibitem{Dong2024}
H.~Dong, Z.~Liu, D.~Hu, X.~Wu, F.~Giorgio-Serchi, and Y.~Yang, ``Tactile sensing on deformed surfaces with electrical impedance tomography,'' in \emph{2024 IEEE International Instrumentation and Measurement Technology Conference (I2MTC)}, pp. 1--6.\hskip 1em plus 0.5em minus 0.4em\relax IEEE, 2024.

\bibitem{Franke1982}
R.~Franke, ``Scattered data interpolation: tests of some methods,'' \emph{Mathematics of computation}, vol.~38, no. 157, pp. 181--200, 1982.

\bibitem{Ioffe2015}
S.~Ioffe and C.~Szegedy, ``Batch normalization: Accelerating deep network training by reducing internal covariate shift,'' in \emph{International conference on machine learning}, pp. 448--456.\hskip 1em plus 0.5em minus 0.4em\relax pmlr, 2015.

\bibitem{Lionheart2004}
W.~R. Lionheart, ``Eit reconstruction algorithms: pitfalls, challenges and recent developments,'' \emph{Physiological measurement}, vol.~25, no.~1, p. 125, 2004.

\bibitem{Kingma2014}
D.~P. Kingma and J.~Ba, ``Adam: A method for stochastic optimization,'' \emph{arXiv preprint arXiv:1412.6980}, 2014.

\bibitem{Vauhkonen1998}
M.~Vauhkonen, D.~Vad{\'a}sz, P.~A. Karjalainen, E.~Somersalo, and J.~P. Kaipio, ``Tikhonov regularization and prior information in electrical impedance tomography,'' \emph{IEEE transactions on medical imaging}, vol.~17, no.~2, pp. 285--293, 1998.

\bibitem{Tehrani2012}
J.~N. Tehrani, A.~McEwan, C.~Jin, and A.~Van~Schaik, ``L1 regularization method in electrical impedance tomography by using the l1-curve (pareto frontier curve),'' \emph{Applied Mathematical Modelling}, vol.~36, no.~3, pp. 1095--1105, 2012.

\bibitem{Liu2019}
S.~Liu, H.~Wu, Y.~Huang, Y.~Yang, and J.~Jia, ``Accelerated structure-aware sparse bayesian learning for three-dimensional electrical impedance tomography,'' \emph{IEEE transactions on industrial informatics}, vol.~15, no.~9, pp. 5033--5041, 2019.

\bibitem{Yang2017}
Y.~Yang and J.~Jia, ``A multi-frequency electrical impedance tomography system for real-time 2d and 3d imaging,'' \emph{Review of Scientific Instruments}, vol.~88, no.~8, 2017.

\bibitem{shooter2005use}
D.~Shooter, ``Use of two-point discrimination as a nerve repair assessment tool: preliminary report,'' \emph{ANZ journal of surgery}, vol.~75, no.~10, pp. 866--868, 2005.

\end{thebibliography}

\vfill

\end{document}